\documentclass{article}  %
\usepackage{iclr2026_conference,times}

\usepackage{amsmath,amsfonts,bm}

\def\eqref#1{equation~\ref{#1}}

\def\1{\bm{1}}

\DeclareMathAlphabet{\mathsfit}{\encodingdefault}{\sfdefault}{m}{sl}
\SetMathAlphabet{\mathsfit}{bold}{\encodingdefault}{\sfdefault}{bx}{n}

\usepackage{capt-of, eucal}
\usepackage{xspace}
\usepackage{xcolor}
\usepackage{enumitem}
\usepackage{amsmath,amsthm,amssymb,amsfonts,dsfont,pifont,bm,bbm,mathrsfs,mathtools,nicefrac,extarrows,relsize}
\usepackage{algorithm,algpseudocode,listings}
\usepackage{booktabs,multirow,adjustbox,diagbox,threeparttable,tabularray,setspace}
\usepackage{wrapfig}
\usepackage{graphicx}
\usepackage{subcaption}
\usepackage{tabularx}
\usepackage{array}
\usepackage{makecell}
\usepackage{longtable} 
\usepackage{paracol}
\usepackage{float}
\usepackage{fontawesome5}
\usepackage[tableposition=top]{caption}
\usepackage{colortbl}
\usepackage{color}
\usepackage{multicol}
\usepackage{lipsum}

\definecolor{citeblue}{rgb}{0.21,0.49,0.74}
\usepackage[pagebackref=false,breaklinks,colorlinks,citecolor=citeblue,bookmarks=false]{hyperref}
\usepackage{wrapfig}
\usepackage[capitalize]{cleveref}  %

\crefname{section}{Sec.}{Secs.}
\Crefname{section}{Section}{Sections}
\crefname{appendix}{Appendix}{Appendices}
\Crefname{appendix}{Appendix}{Appendices}
\crefname{table}{Table}{Tables}
\Crefname{table}{Table}{Tables}
\crefname{figure}{Fig.}{Figs.}
\Crefname{figure}{Figure}{Figures}
\crefname{equation}{Eq.}{Eqs.}
\Crefname{equation}{Equation}{Equations}
\crefname{theorem}{Thm.}{Thms.}
\Crefname{theorem}{Theorem}{Theorems}
\crefname{lemma}{Lem.}{Lems.}
\Crefname{lemma}{Lemma}{Lemmas}
\crefname{remark}{Rem.}{Rems.}
\Crefname{remark}{Remark}{Remarks}
\crefname{corollary}{Cor.}{Cors.}
\Crefname{corollary}{Corollary}{Corollaries}
\crefname{algorithm}{Alg.}{Algs.}
\Crefname{algorithm}{Algorithm}{Algorithms}
\definecolor{cellred}{RGB}{213, 123, 101}
\definecolor{cellgreen}{RGB}{0, 205, 0}
\definecolor{cellblue}{RGB}{54, 125, 189}
\definecolor{codegreen}{rgb}{0,0.6,0}
\definecolor{codegray}{rgb}{0.5,0.5,0.5}
\definecolor{codepurple}{rgb}{0.58,0,0.82}
\definecolor{backcolour}{rgb}{1.0,1.0,1.0}
\lstdefinestyle{mystyle}{
    backgroundcolor=\color{backcolour},
    commentstyle=\color{codegreen},
    keywordstyle=\color{magenta},
    numberstyle=\tiny\color{codegray},
    stringstyle=\color{codepurple},
    basicstyle=\ttfamily\scriptsize,
    breakatwhitespace=false,
    breaklines=true,
    captionpos=b,
    keepspaces=true,
    numbers=left,
    numbersep=5pt,
    showspaces=false,
    showstringspaces=false,
    showtabs=false,
    tabsize=2
}
\usepackage[most,skins,theorems]{tcolorbox}
\tcbset{
  aibox/.style={
    width=\linewidth,
    top=8pt,
    bottom=4pt,
    colback=blue!6!white,
    colframe=black,
    colbacktitle=black,
    enhanced,
    center,
    attach boxed title to top left={yshift=-0.1in,xshift=0.15in},
    boxed title style={boxrule=0pt,colframe=white,},
  }
}
\newtcolorbox{AIbox}[2][]{aibox,title=#2,#1}
\usepackage{amsmath} %
\usepackage{amssymb} %
\usepackage{multirow}

\usepackage{rotating}  %

\newcolumntype{C}[1]{>{\centering\arraybackslash}p{#1}}
\newcolumntype{L}[1]{>{\arraybackslash}p{#1}}

\definecolor{demphcolor}{gray}{.2}

\definecolor{demphcolorinline}{gray}{.3}

\definecolor{demphcolor1}{gray}{.6}

\newcommand{\tocite}[1]{{\color{red} [TO CITE]}}

\newcommand{\methodname}{{HBPO}\xspace}

\title{Hierarchical Budget Policy Optimization for\\ Adaptive Reasoning}

\author{%
  \textbf{Shangke Lyu}$^{1,*}$, 
  ~~
  \textbf{Linjuan Wu}$^{1,*}$,
  ~~
  \textbf{Yuchen Yan}$^{1}$,
  ~~
  \textbf{Xingyu Wu}$^{1}$,
  ~~
  \textbf{Hao Li}$^{2}$ 
  \\
  \textbf{Yongliang Shen}$^{1}$,
  ~~
  \textbf{Peisheng Jiang}$^{2}$,
  ~~
  \textbf{Weiming Lu}$^{1}$,
  ~~
  \textbf{Jun Xiao}$^{1}$,
  ~~
  \textbf{Yueting Zhuang}$^{1}$ \\
  $^1$Zhejiang University \quad $^2$SF Technology \\
  \texttt{\{lyusk, wulinjuan525, syl, luwm\}@zju.edu.cn} \\
  \vspace{0.1cm} \\
  \begin{tabular}{@{}ll@{}}
    \faGithub\ GitHub: & \href{https://github.com/zju-real/hbpo}{\texttt{\textcolor{cyan}{https://github.com/zju-real/hbpo}}} \\
    \faGlobe\ Project: & \href{https://zju-real.github.io/hbpo/}{\texttt{\textcolor{cyan}{https://zju-real.github.io/hbpo}}}
  \end{tabular}
}

\iclrfinalcopy %

\begin{document}

\maketitle

\renewcommand{\thefootnote}{\fnsymbol{footnote}}
\footnotetext[1]{~The first two authors have equal contributions.}
\renewcommand{\thefootnote}{\arabic{footnote}}

\begin{abstract}

Large reasoning models achieve remarkable performance through extensive chain-of-thought generation, yet they suffer from a critical inefficiency: applying uniformly extensive reasoning regardless of problem complexity.  We present Hierarchical Budget Policy Optimization (\methodname), a reinforcement learning framework that enables models to learn problem-specific reasoning depths without sacrificing capability.
Unlike existing approaches that impose rigid constraints or rely on discrete mode selection, \methodname partitions the exploration space into budget-constrained hierarchies (512-2560 tokens), each with differentiated reward structures that preserve both efficiency incentives and reasoning capabilities.
This design addresses a fundamental challenge in efficient reasoning training: traditional length penalties systematically bias models away from necessary long reasoning paths, causing exploration space collapse.
Through hierarchical sampling and budget-aware rewards, \methodname maintains exploration diversity while teaching models to recognize when extended deliberation is warranted.
Extensive experiments demonstrate that \methodname reduces average token usage by up to 60.6\% while improving accuracy by 3.14\% across four reasoning benchmarks. 
Most notably, \methodname exhibits emergent adaptive behavior where models automatically adjust reasoning depth based on problem complexity. Our results suggest that reasoning efficiency and capability are not inherently conflicting, and can be simultaneously optimized through appropriately structured hierarchical training that preserves exploration diversity.

\end{abstract}

\section{Introduction}

\begin{figure}[t]
  \centering
  \includegraphics[width=0.98\textwidth]{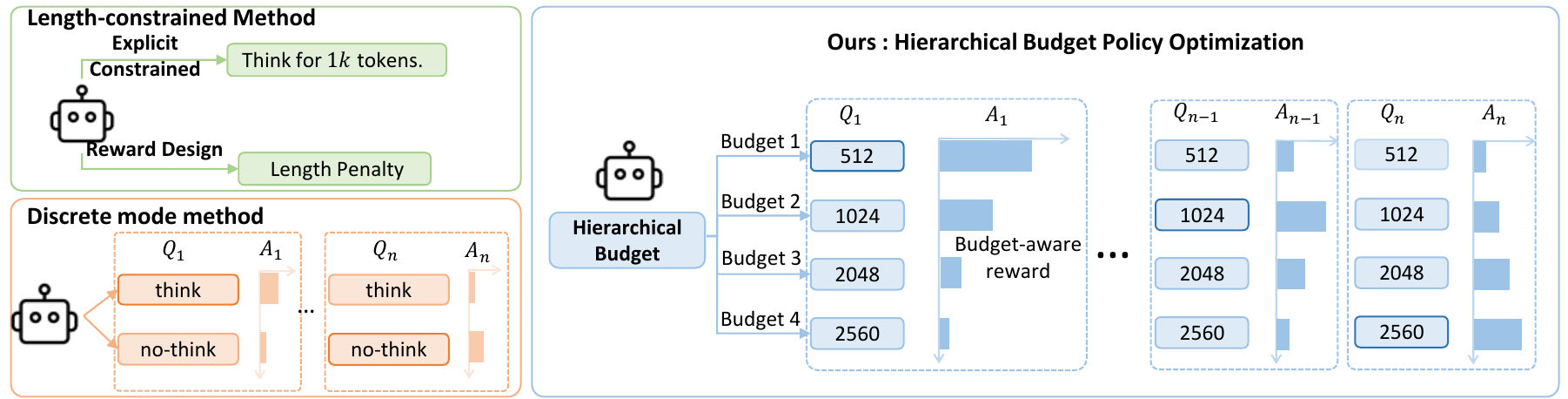}
  \caption{HBPO provides budget-aware reward through hierarchical budget exploration, which enables fine-grained adaptive reasoning. While length-constrained methods use global constraint or length penalty, and discrete mode methods dichotomize problem difficulty, HBPO partitions the exploration space into budget-constrained hierarchies (512, 1024, 2048, 2560 tokens). This structure maintains reasoning diversity throughout training, enabling emergent adaptive behavior where models match computational resources to problem complexity.}
  \label{fig:1}
\end{figure}

Advances in large reasoning models have led to impressive performance on complex reasoning tasks through chain-of-thought methodologies~\citep{LearningReasonLLMs,guoDeepseekr1IncentivizingReasoning2025}. However, these models exhibit fundamental inefficiency: they generate unnecessarily long reasoning chains even for simple problems, sometimes consuming thousands of tokens for basic arithmetic~\citep{chenReasoningEraSurvey2025,chenNOTThinkThat2025}. This phenomenon reveals a fundamental misalignment, as current reasoning models lack the ability to adapt their computational effort to the actual complexity of problems.

Recent empirical findings challenge the conventional belief that longer reasoning always leads to better outcomes. Research shows that models can maintain competitive accuracy even without intermediate steps~\citep{maReasoningModelsCan2025}, and in some cases, shorter reasoning paths perform comparably or even better on simpler tasks~\citep{liSelfBudgeterAdaptiveToken2025}. This is further supported by stark variations in optimal reasoning lengths across tasks. For instance, L1~\citep{aggarwalL1ControllingHow2025} achieves peak performance with $\sim$1,100 tokens on GSM8K, but requires over 3,000 tokens on OlympiadBench. Such heterogeneity highlights a key insight: the computational requirements for effective reasoning are inherently problem-dependent, yet current models apply uniform reasoning strategies regardless of task complexity.

To address these inefficiencies, an increasing number of studies aim to improve the inference efficiency of reasoning models. Current approaches fall into two primary categories. \textbf{\textit{Length-constrained methods}} directly constrain generation through explicit mechanisms or incorporate length penalties into training objectives: prompts like ``think for n tokens" and corresponding length-control rewards in L1~\citep{aggarwalL1ControllingHow2025}; progressively limits on the model's reasoning space during training in ThinkPrune~\citep{houThinkPrunePruningLong2025}; enforces budget constraints through forced termination in Scalable Chain of Thoughts~\citep{xuScalableChainThoughts2025}; and HAPO~\citep{huangHAPOTrainingLanguage2025} leverages history-aware optimization to track minimal sufficient reasoning lengths. \textbf{\textit{Discrete mode methods}} dichotomize problem difficulty and omit the reasoning process for simple instances, which enables the model to operate in a think/no-think manner. Thinkless~\citep{fangThinklessLLMLearns2025} first performs format training for mode switching via fine-tuning. AdaptThink~\citep{zhangAdaptThinkReasoningModels2025} employs importance sampling to enable the model to switch between reasoning patterns. While effective at reducing token usage, these methods share a key limitation: they prioritize efficiency or mode selection at the cost of accuracy performance, lacking fine-grained mechanisms for models to autonomously decide appropriately efficient reasoning length.

We identify two key challenges that hinder existing methods from achieving genuine reasoning efficiency. \textbf{First, length penalties introduce systematic training biases that impair reasoning capabilities.} In standard reinforcement learning settings~\citep{guoDeepseekr1IncentivizingReasoning2025}, correct solutions receive equal rewards regardless of length, allowing for unbiased exploration. However, length penalties disrupt this balance by consistently favoring shorter outputs, leading policies to gradually abandon long-reasoning strategies~\citep{houThinkPrunePruningLong2025,huangHAPOTrainingLanguage2025,louAdaCoTParetoOptimalAdaptive2025}.
\textbf{Second, static efficiency constraints fail to capture the continuous nature of reasoning complexity.} Even adaptive methods rely on coarse mechanisms, such as binary think/no-think decisions~\citep{zhangAdaptThinkReasoningModels2025,fangThinklessLLMLearns2025} or fixed confidence thresholds~\citep{qiaoConCISEConfidenceguidedCompression2025}, which overlook the nuanced relationship between problem characteristics and computational requirements.

These limitations raise a fundamental question: \textbf{\textit{rather than enforcing uniform constraints, can models learn differentiated reasoning strategies through structured exploration?}} This question motivates our study of hierarchical budget exploration, where efficiency emerges not from rigid control but from structured exploration within budget-constrained subspaces.

We propose \textbf{Hierarchical Budget Policy Optimization (\methodname)} illustrated in Figure~\ref{fig:1}, a reinforcement learning framework that enables models to learn problem-specific reasoning strategies while retaining their ability to perform complex reasoning. The core idea is to partition the exploration space into multiple budget-constrained subgroups, allowing models to preserve reasoning diversity and uncover natural alignments between problem characteristics and required computational effort. 
Specifically, \methodname employs a hierarchical sampling strategy that partitions rollout samples into subgroups, each governed by a distinct token budgets. We implement this by inserting length prompts (e.g., \textit{``I will answer the question within n tokens"}) after the reasoning tag, thereby constructing multiple exploration spaces with budgets ranging from 512 to 2560 tokens. Unlike uniform sampling, this structure encourages the model to explore both concise and extended reasoning paths throughout training, effectively mitigating the systematic degradation of reasoning capabilities caused by global length penalties.

To enable efficient reasoning within each budget hierarchy, we design a piecewise reward function with distinct behaviors inside and outside budget boundaries. 
Within the assigned budget, rewards are monotonically non-decreasing to preserve exploratory flexibility. Beyond the budget, cosine decay and length deviation penalties are applied to encourage the model to return to its designated exploration space. This creates a natural gradient of incentives: shorter budgets favor concise solutions with higher rewards, while longer budgets retain standard rewards for extended reasoning.
The result is a reward landscape that teaches models not just to reason efficiently within constraints, but to recognize which constraint level matches the problem at hand.

\methodname achieves a superior accuracy-efficiency trade-off compared to existing methods on four reasoning benchmarks. Crucially, it exhibits adaptive behavior by dynamically allocating computational resources based on problem complexity. For example, on GSM8K, it uses only 670 tokens. On AIME25, it uses 5,606 tokens, representing a more than eightfold increase in token usage. In both cases, it improves accuracy by 2.2\% and 8.9\% compared to the base model DeepSeek-R1-Distill-Qwen-1.5B, demonstrating effective resource allocation.

Our contributions are threefold:

\begin{itemize}
\item We introduce Hierarchical Budget Policy Optimization, a reinforcement learning framework that partitions the exploration space into budget-constrained hierarchies with differentiated rewards, preserving reasoning diversity while enabling adaptive resource allocation.
\item We demonstrate that uniform efficiency constraints systematically collapse the exploration space and degrade reasoning capabilities, validating the necessity of structured exploration for maintaining model performance.
\item We provide evidence of emergent adaptive reasoning, where HBPO-trained models automatically adjust reasoning depth based on problem characteristics, achieving up to 60.6\% reduction in token usage while improving accuracy by 3.14\% across mathematical reasoning benchmarks.
\end{itemize}

\section{Related Works}
\subsection{Efficient Reasoning}

Recent advances in reasoning models have spurred various efforts to reduce computational overhead while preserving performance. Existing approaches can be broadly categorized into three types: Length-constrained methods explicitly restrict generation through predefined mechanisms. For example, L1~\citep{aggarwalL1ControllingHow2025} introduces token budget prompts with corresponding rewards; ThinkPrune~\citep{houThinkPrunePruningLong2025} progressively tightens constraints via iterative training; and Scalable Chain of Thoughts~\citep{xuScalableChainThoughts2025} separates the thinking and solution phases, each with its budget. While effective in limiting token usage, these methods require manual budget specification and lack adaptability to varying problem complexity. Reward-based methods incorporate efficiency into training objectives more implicitly. HAPO~\citep{huangHAPOTrainingLanguage2025} incentivizes concise reasoning by tracking minimal correct response lengths, while “Think When You Need”~\citep{jiangThinkOnlyWhen2025} balances brevity and quality through pairwise comparisons and adaptive target lengths. These approaches offer finer control but still impose global objectives across diverse problem types, limiting flexibility. Training-free approaches~\citep{muennighoffS1SimpleTesttime2025,yangDynamicEarlyExit2025} intervene at inference time through symbolic control tokens or confidence-based early stopping. While cost-effective, these methods are heuristic-driven and lack learning-based adaptation. Despite their differences, all these approaches share a fundamental limitation: they treat efficiency as a uniform constraint, overlooking the fact that optimal reasoning length varies significantly with problem complexity.

\subsection{Adaptive Reasoning}

Recognizing heterogeneous reasoning requirements, recent work explores adaptive strategies that adjust computational effort based on problem characteristics. Binary mode selection represents the most common approach, with models choosing between thinking and non-thinking modes~\citep{louAdaCoTParetoOptimalAdaptive2025,zhangAdaptThinkReasoningModels2025,fangThinklessLLMLearns2025}. These methods employ various techniques including selective loss masking, simplified mode definitions, and decoupled optimization to prevent mode collapse. Multi-stage training strategies~\citep{jiangThinkOnlyWhen2025,tuLearningWhenThink2025,zhangWhenContinueThinking2025} use sophisticated reward designs and batch-level balancing to achieve better mode distributions. Beyond binary selection, multi-modal approaches define richer reasoning taxonomies: ARM~\citep{wuARMAdaptiveReasoning2025} uses four modes with adaptive scaling, while PATS~\citep{wangPATSProcessLevelAdaptive2025} enables step-level switching between complexity levels. Some methods introduce auxiliary components like regression models for mode prediction~\citep{liangThinkSwitcherWhenThink2025} or self-budgeting mechanisms~\citep{liSelfBudgeterAdaptiveToken2025}. While these adaptive approaches demonstrate significant efficiency gains, they operate within discrete categories rather than enabling continuous adaptation. Complex multi-stage procedures and predefined mode taxonomies limit their flexibility and generalization. In contrast, our hierarchical budget exploration framework enables continuous adaptation through a unified policy optimization process. Without relying on manually defined modes or external modules, our approach allows the model to learn problem-specific reasoning depths, leading to emergent adaptive behavior that naturally aligns computational effort with problem complexity.

\section{Method}

\begin{figure}[t]
\centering
\includegraphics[width=0.98\textwidth]{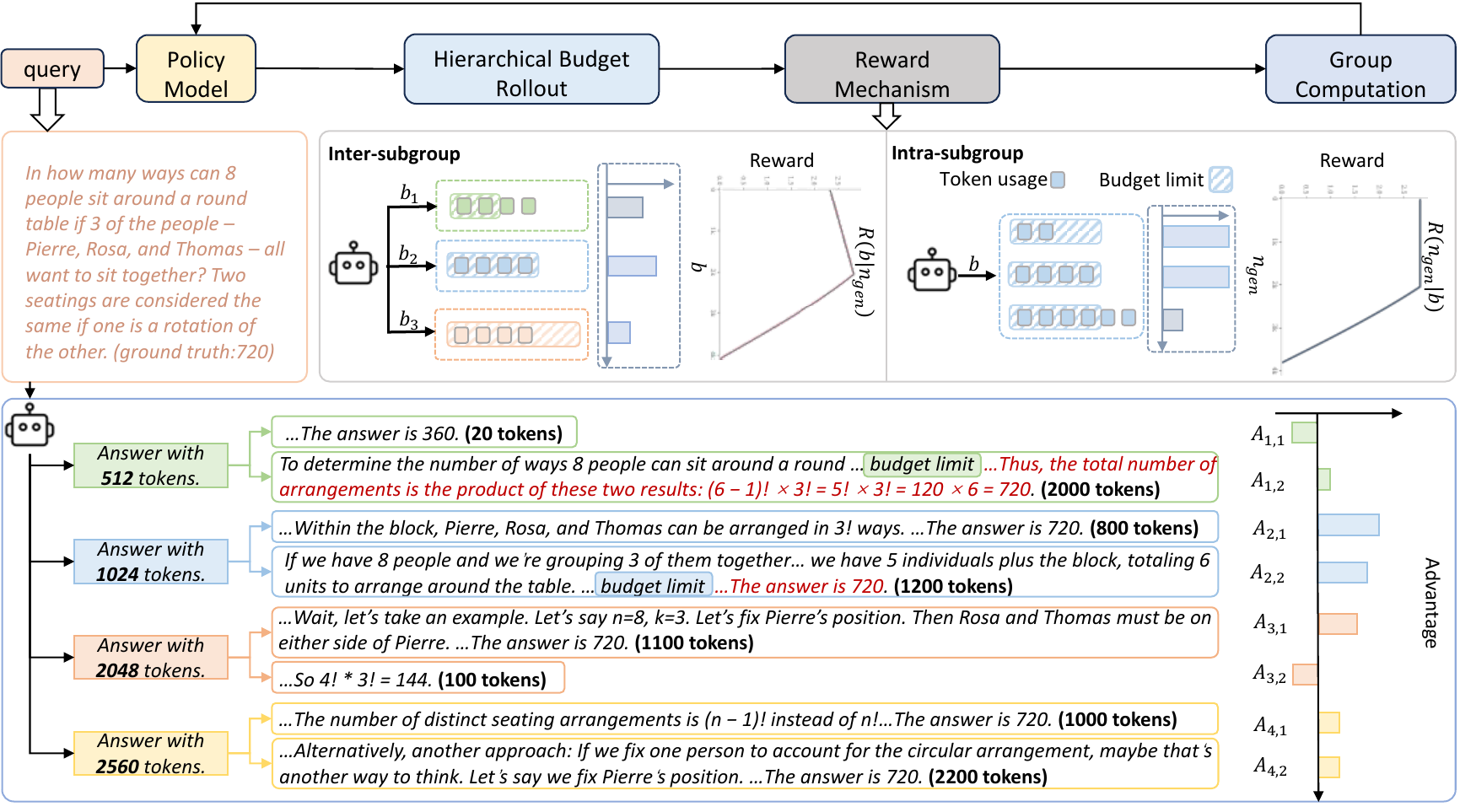}
\caption{Overview of Hierarchical Budget Policy Optimization. Given a query, HBPO generates responses across multiple budget-constrained subgroups (512, 1024, 2048, 2560 tokens), each guided by a piecewise reward function that preserves exploration within budgets while penalizing excess through deviation penalties. The advantage computation decomposes into intra-subgroup advantages (comparing responses against budget-specific baselines) and inter-subgroup advantages (enabling cross-budget learning through global comparison). This hierarchical structure enables models to learn efficient reasoning within constraints and adaptive budget selection based on problem complexity.}
\label{fig:hbpo_overview}
\end{figure}

We present Hierarchical Budget Policy Optimization, as shown in Figure~\ref{fig:hbpo_overview}, which extends the Group Relative Policy Optimization (GRPO)~\citep{guoDeepseekr1IncentivizingReasoning2025} framework to enable adaptive reasoning through structured exploration. The core innovation lies in partitioning the exploration space into budget-constrained hierarchies and designing differentiated reward mechanisms that preserve reasoning diversity. We first introduce the hierarchical rollout strategy (Section \ref{sec:3.1}), then detail the budget-aware reward design (Section \ref{sec:3.2})), and finally describe the training procedure (Section \ref{sec:3.3})).

\subsection{Hierarchical Budget Exploration}

\label{sec:3.1}

The fundamental challenge in efficient reasoning training is that uniform length penalties systematically bias models away from necessary long reasoning paths. To address this, we partition rollout samples into hierarchical subgroups, each operating within distinct token budget constraints. This structure ensures that models maintain exposure to diverse reasoning lengths throughout training.

Given a query $q$, we generate $n$ rollout samples and partition them into $k$ subgroups $\{G_1, G_2, ..., G_k\}$, where each subgroup $G_i$ is associated with a token budget $b_i$. We implement this through budget-specific prompts embedded after the reasoning tag: "I will answer the question within $b_i$ tokens". The budget values form an ascending sequence $(b_1 < b_2 < ... < b_k)$, spanning from compact reasoning (e.g., 512 tokens) to extended deliberation (e.g., 2560 tokens).

This hierarchical structure serves two key purposes. First, it prevents exploration space collapse, a common issue in efficiency training where models abandon long reasoning. By preserving separate exploration spaces, \methodname ensures sampling across diverse reasoning lengths. Second, it enables structured comparative learning: the model discovers the suitable computation for each problem by contrasting performance across budget levels, rather than relying on global optimization.

\subsection{Budget-Aware Reward Design}
\label{sec:3.2}

The effectiveness of hierarchical exploration hinges on careful reward design. Existing methods either use uniform rewards—supporting fair exploration but lacking efficiency incentives—or apply global length penalties, which improve efficiency at the cost of reasoning ability. HBPO addresses this trade-off with a piecewise reward function that integrates the strengths of both approaches.

\subsubsection{Intra-Budget Reward Function}

Within each budget-constrained subgroup, we design a reward function that balances reason exploration and efficiency. For a given budget $b$, the reward integrates length-based penalties $f_1$ that promote token efficiency with classical rewards $f_2$  that encourage diverse reasoning. The reward is formally defined as:
\begin{equation}
R(n_{\text{gen}} \mid b) = \begin{cases} 
f_1(n_{\text{gen}}, b), & \text{if correct, } n_{\text{gen}} > b, \text{ and } n_{\text{gen}} \leq L_{\max} \\
f_2(b), & \text{if correct, } n_{\text{gen}} \leq b, \text{ and } n_{\text{gen}} \leq L_{\max} \\

0, & \text{otherwise}
\end{cases}
\label{eq:reward_intra}
\end{equation}
where:
\begin{align}
f_1(n_{\text{gen}}, b) &= \beta \cdot \cos\left(\frac{\pi n_{\text{gen}}}{2L_{\max}}\right) - \alpha |n_{\text{gen}} - b| \label{eq:f1} \\
f_2(b) &= \beta \cdot \cos\left(\frac{\pi b}{2L_{\max}}\right) \label{eq:f2}
\end{align}
Here, $n_{\text{gen}}$ denotes the number of generated tokens, $L_{\max}$ is the maximum context length, $\beta$ is a scaling factor, and $\alpha$ controls deviation sensitivity. The piecewise structure serves distinct purposes across different generation lengths. When $n_{\text{gen}} > b$, the reward follows $f_1$, incorporating both cosine decay and deviation penalty to guide the model back to its designated exploration space. When $n_{\text{gen}} \leq b$, the reward is bounded by $f_2$, ensuring monotonic non-decreasing behavior that preserves exploration within the budget.

\subsubsection{Inter-Budget Reward Differentiation}

The hierarchical structure naturally induces reward differentiation across budgets. For a fixed generation length $n_{\text{gen}}$, different budget assignments yield different rewards according to Equation \ref{eq:reward_intra}, signaled as $R(b \mid n_{\text{gen}})$. This creates systematic preferences that align with problem complexity.

When $n_{\text{gen}} < \min(b_i)$, all budgets yield rewards determined by $f_2$, and smaller budgets receive higher rewards due to the monotonic decrease of the cosine function over the interval. This preference for smaller budgets on short responses encourages efficiency for simple problems. Conversely, when $n_{\text{gen}} > \max(b_i)$, larger budgets provide higher rewards through smaller deviation penalties $|n_{\text{gen}} - b_i|$ in $f_1$, preserving the model's ability to engage in extended reasoning when necessary.

 As $n_{\text{gen}}$ increases from below $\min(b_i)$ to above $\max(b_i)$, the reward functions corresponding to different budgets transition in relative preference. The intersection points between reward curves represent complexity thresholds where the optimal budget choice transitions. Through comparative advantage across these differentiated rewards, the model learns to match computational resources to problem requirements without explicit complexity labels or external guidance.

\begin{algorithm}[t]
\caption{Hierarchical Budget Policy Optimization (HBPO)}
\label{alg:hbpo}
\begin{algorithmic}[1]
\Require Initial policy $\pi_{\theta_0}$, budget levels $\mathcal{B} = \{b_1, ..., b_k\}$, learning rate $\eta$
\For{iteration $t = 1, 2, ..., T$}
    \State Sample batch of queries $\mathcal{Q}$ from training data
    \For{each query $q \in \mathcal{Q}$}
        \For{each budget $b_i \in \mathcal{B}$}
            \State Generate $n/k$ responses with prompt ``I will answer within $b_i$ tokens''
            \State Store responses in subgroup $G_i$
        \EndFor
        \For{each subgroup $G_i$}
            \State Compute rewards $\{R_{i,j}\}$ using Equation \ref{eq:reward_intra}
             \State Compute intra-subgroup mean reward: $\mu_i = \frac{1}{|G_i|} \sum_{j=1}^{|G_i|} R_{i,j}$
            \State Compute budget rewards $R_{b_i}$ using Equation \ref{eq:f2}
             \State Compute intra-subgroup advantage: $A_{i}^{\text{intra}} = \mu_i - R_{b_i}$
        \EndFor
        \State Compute inter-subgroup advantage: $A_{i,j}^{\text{inter}} = \frac{R_{i,j} - \frac{1}{n} \sum_{i,j} R_{i,j}}{\mathrm{std}(R)}$
            \State Normalize final advantage: $A_{i,j} = A_{i}^{\text{intra}} + A_{i,j}^{\text{inter}}$
        
    \EndFor
    \State Update policy: $\theta_{t+1} \leftarrow \theta_t - \eta \nabla_\theta \mathcal{L}(\theta_t)$
\EndFor
\end{algorithmic}
\end{algorithm}

\subsection{Training Procedure}
\label{sec:3.3}

HBPO extends the standard GRPO framework by incorporating hierarchical sampling and budget-aware advantage computation into the policy optimization process, the algorithm is shown in Algorithm~\ref{alg:hbpo}. During each training iteration $t$, the model generates $n$ responses for a given query, which are automatically partitioned into $k$ subgroups based on their associated budget constraints. Each response is generated with an embedded budget prompt "I will answer the question within $b_i$ tokens", where $b_i \in \{b_1, b_2, ..., b_k\}$ represents the predetermined budget levels.

The advantage computation leverages the hierarchical structure to enable both efficient reasoning within budgets and adaptive budget selection across problems. For the $j$-th response in the $i$-th subgroup, we compute the reward $R_{i,j}$ using the budget-aware reward function described in Section \ref{sec:3.2}. To capture the hierarchical nature of our exploration, we decompose the advantage into two complementary components that guide different aspects of learning.

The intra-subgroup advantage measures how well responses perform relative to their budget expectation:
$A_{i}^{\text{intra}} = \mu_i - R_{b_i}$, where $\mu_i = \frac{1}{|G_i|} \sum_{j=1}^{|G_i|} R_{i,j}$ is the mean reward within subgroup $i$, and $R_{b_i}$ represents the budget-specific baseline computed using Equation~\ref{eq:f2}. This term encourages optimization within each budget constraint, teaching the model to reason efficiently given a specific token allocation.

The inter-subgroup advantage enables comparative learning across different budgets:
\begin{equation}
A_{i,j}^{\text{inter}} = \frac{R_{i,j} - \frac{1}{n} \sum_{i,j} R_{i,j}}{\text{std}(R)}
\end{equation}
This term compares each response against the global mean, creating natural preferences for budget selection. Responses from shorter budgets that achieve high rewards receive positive advantages, while unnecessarily long responses receive negative advantages, teaching the model to match computational effort to problem requirements.

The final advantage combines both components with normalization for stable training:
\begin{equation}
A_{i,j} = A_{i}^{\text{intra}} + A_{i,j}^{\text{inter}}
\end{equation}
The policy optimization adopts GRPO's clipped objective to prevent destructive updates:
\begin{equation}
\mathcal{L}(\theta) = -\mathbb{E}_{(s,a) \sim \pi_{\theta_{\text{old}}}}\left[\min\left(\rho_\theta(s,a)A(s,a), \operatorname{clip}(\rho_\theta(s,a), 1-\epsilon_{\text{low}}, 1+\epsilon_{\text{high}})A(s,a)\right)\right]
\end{equation}
where $\rho_\theta(s,a) = \pi_\theta(a|s) / \pi_{\theta_{\text{old}}}(a|s)$ represents the probability ratio. The hierarchical advantages $A_{i,j}$ naturally flow through this objective, enabling the model to improve both within-budget efficiency and cross-budget selection without requiring separate optimization objectives or complex multi-stage training procedures.

\section{Experiments}
\subsection{Experimental Setup}

\paragraph{Datasets and Models.}

We evaluate HBPO on mathematical reasoning tasks using the DeepScaleR dataset~\citep{deepscaler2025} for training, which comprises 40K high-quality mathematical problems from AIME, AMC, Omni-Math~\citep{gao2024omnimathuniversalolympiadlevel}, and STILL~\citep{minImitateExploreSelfImprove}. We employ two base models: DeepSeek-R1-Distill-Qwen-1.5B~\citep{guoDeepseekr1IncentivizingReasoning2025} and DeepScaleR-Preview-1.5B~\citep{deepscaler2025}.

\paragraph{Implementation Details.} We implement HBPO using the VeRL framework~\citep{sheng2024hybridflow} with a context window of 4,096 tokens during training. Following DAPO~\citep{yuDAPOOpenSourceLLM2025a}, we set clipping thresholds $\epsilon_{\text{high}} = 0.28$ and $\epsilon_{\text{low}} = 0.2$, with KL divergence disabled to encourage exploration. Training proceeds for one epoch (629 steps) with a learning rate of $10^{-6}$ and batch size of 64. For hierarchical exploration, we generate 16 rollouts per query, partitioned equally into 4 subgroups with budget constraints $\mathcal{B} = {512, 1024, 2048, 2560}$ tokens.

\paragraph{Evaluation Protocol.}
We evaluate on four mathematical reasoning benchmarks of increasing difficulty: GSM8K~\citep{cobbe2021gsm8k}, Math500~\citep{lightman2023lets}, OlympiadBench~\citep{he2024olympiadbench}, and AIME25. Following standard practice~\citep{guoDeepseekr1IncentivizingReasoning2025}, we use temperature $T=0.6$, $top\_p=0.95$, and maximum context length of 32,768 tokens. We report pass@1 accuracy and average token usage under two evaluation settings: (1) natural reasoning where models freely determine their computational effort, and (2) efficiency prompting using \textit{``I will answer the question with minimal tokens"} after \texttt{
<think>} to guide models toward efficient responses.

\paragraph{Baselines.}

We compare against several state-of-the-art efficient reasoning methods: (1) global penalties: HAPO ~\citep{huangHAPOTrainingLanguage2025} and TLMRE~\citep{aroraTrainingLanguageModels2025} add length penalties to the RL objective; (2) explicit control: L1-Exact,L1-Max~\citep{aggarwalL1ControllingHow2025}, E1~\citep{xuScalableChainThoughts2025} and ThinkPrune~\citep{houThinkPrunePruningLong2025}use RL with explicit length targets. (3) discrete mode selection: AdaptThink~\citep{zhangAdaptThinkReasoningModels2025}, AutoThink~\citep{tuLearningWhenThink2025} AdaR1~\citep{luoAdaR1LongCoTHybridCoT2025} and Thinkless~\citep{fangThinklessLLMLearns2025} enable binary think/no-think mode selection.

\subsection{Main Results}

\paragraph{Hierarchical training enables efficient reasoning without capability trade-offs.}

Tables~\ref{tab:natural_reasoning} and \ref{tab:constrained_reasoning} present our results under natural and efficiency-constrained settings, respectively. Under natural reasoning conditions, HBPO demonstrates consistent improvements across both base models. Applied to DeepSeek-R1-Distill-Qwen-1.5B, HBPO improves average accuracy from 56.3\% to 59.4\% while reducing token usage by 60.6\% (from 7,921 to 3,120). On the stronger DeepScaleR model, HBPO maintains the baseline's 63.7\% accuracy while achieving 50.2\% token reduction (from 4,744 to 2,364). Notably, HBPO achieves 31.1\% accuracy on AIME25, outperforming the DeepScaleR baseline and all efficiency methods. This improvement on the most challenging benchmark while using fewer tokens demonstrates that hierarchical exploration not only prevents capability degradation but can enhance reasoning by eliminating computational redundancy.

The efficiency prompting setting makes the performance gains from hierarchical training more evident. While baseline models suffer catastrophic degradation when forced to minimize tokens (over 10\% accuracy drop), HBPO maintains robust performance. Applied to DeepScaleR, HBPO achieves 59.4\% average accuracy with only 947 tokens, matching L1-Max (1024)'s accuracy while using 32\% fewer tokens. This indicates that our training enables effective exploration across the entire efficiency spectrum.

\begin{table*}[h!]
\small
\centering
\small
\begin{tabular}{lcccccccc>{\columncolor{gray!10}}c>{\columncolor{gray!10}}c}
\toprule
\multirow{2}{*}{\textbf{Method}} & \multicolumn{2}{c}{\textbf{GSM8K}} & \multicolumn{2}{c}{\textbf{Math500}} & \multicolumn{2}{c}{\textbf{Olympiad}} & \multicolumn{2}{c}{\textbf{AIME25}} & \multicolumn{2}{>{\columncolor{gray!10}}c}{\textbf{Average}} \\
\cmidrule(lr){2-3} \cmidrule(lr) {4-5} \cmidrule(lr){6-7}\cmidrule(lr){8-9} \cmidrule(lr){10-11} 
& Acc & Tokens & Acc & Tokens & Acc & Tokens & Acc & Tokens & Acc & Tokens \\
\midrule
\multicolumn{11}{c}{\textit{Base: DeepSeek-R1-Distill-Qwen-1.5B}} \\
\midrule
Baseline & 82.3 & 1,111 & \textbf{81.6} & 4,696 & 42.3 & 10,225 & 18.9 & 15,651 & 56.3 & 7,921 \\
HAPO &80.9 & 571& 76.4& 2,252& 42.1&5396& \underline{24.4}& 9,230&56.0 & 4362\\
TLMRE & 74.6 & 221 & 69.8 & 1,835 & 35.8 & 4,838 & 17.8 & 9,753 & 49.5 & 4,162 \\
AdaptThink & \textbf{85.0} & 816 & 79.6 & 1,220 &\underline{42.9} & 2,501 & 18.9 & 6,813 & 56.6 & \textbf{2,838} \\
AutoThink &81.4 &739 & 81.4& 2627&44.5 &5709 & 23.3& 9,769&\underline{57.7} & 4,711\\
AdaR1 &79.2 & 341& \underline{80.8} & 2,455&42.1 &5,802 & 23.0& 9,516&56.3 &4,528 \\

\textbf{HBPO (Ours)} & \underline{84.5} & 670 & 80.4 & 2,147 & \textbf{45.0} & 4,058 & \textbf{27.8} & 5,606 & \textbf{59.4} & \underline{3,120} \\
\midrule
\multicolumn{11}{c}{\textit{Base: DeepScaleR-Preview-1.5B}} \\
\midrule
Baseline & 86.1 & 1,684 & \textbf{87.0} & 2,938 & \textbf{51.6} & 5,330 & 30.0 & 9,023 & \underline{63.7} & 4,744 \\
HAPO & 84.3& 658 &84.4 &2,102 & 47.7&3,569 &26.7 & 5,353& 60.8&2,920 \\
ThinkPrune  & \underline{86.6} & 659 & 85.2 & 1,757 & 50.6 & 3,122 & 26.7 & 4,816 & 62.3 & 2,589\\
L1-Exact & 86.4&861 &80.8 & 3685&46.0 &3,478 &23.3 &3,285&59.1&2,827 \\
L1-Max & 86.1 & 670 & 85.0 & 3,260 & 48.2 & 3,094 & 22.2 & 3,163 & 60.4 & \underline{2,547} \\
E1 & 85.4& 748&84.8 &1,930 & 49.3&3,456 &26.7 & 5,729&61.6 &2,965 \\
AutoThink & 85.8 & 1,171 & 81.0 & 2154 & 48.2 & 4,501 & \underline{30.0} & 7,435 & 61.3 & 3,815 \\

Thinkless   & 86.4 & 957 & 85.2 & 3,184 & \underline{50.7} & 5,691 & 25.6 & 8,271 & 62.0 & 4,526\\
\textbf{HBPO (Ours)} & \textbf{87.6} & 790 & \underline{86.2} & 1,818 & 50.0 & 2,861 & \textbf{31.1} & 3,988 & \textbf{63.7} & \textbf{2,364} \\
\bottomrule
\end{tabular}
\caption{Performance under natural reasoning setting. \textbf{Bold} indicates the best and \underline{underline} indicates the second-best for each metric. HBPO achieves the best performance in terms of the accuracy-efficiency trade-off and exhibits adaptive behavior.}
\label{tab:natural_reasoning}
\end{table*}

\begin{table*}[h!]
\centering
\small
\begin{tabular}{lcccccccc>{\columncolor{gray!10}}c>{\columncolor{gray!10}}c}
\toprule
\multirow{2}{*}{\textbf{Method}} & \multicolumn{2}{c}{\textbf{GSM8K}} & \multicolumn{2}{c}{\textbf{Math500}} & \multicolumn{2}{c}{\textbf{Olympiad}} & \multicolumn{2}{c}{\textbf{AIME25}} & \multicolumn{2}{>{\columncolor{gray!10}}c}{\textbf{Average}} \\
\cmidrule(lr){2-3} \cmidrule(lr){4-5} \cmidrule(lr){6-7}\cmidrule(lr){8-9} \cmidrule(lr){10-11} 
& Acc & Tokens & Acc & Tokens & Acc & Tokens & Acc & Tokens & Acc & Tokens \\
\midrule
\multicolumn{11}{c}{\textit{Base: DeepSeek-R1-Distill-Qwen-1.5B}} \\
\midrule
Baseline & \underline{73.6} & 267 & \underline{67.4} & 806 & \underline{30.6} & 1,950 & \underline{13.3} & 3,737 & \underline{46.2} & \underline{1,690} \\
\textbf{HBPO (Ours)} & \textbf{83.9} & 340 & \textbf{79.6} & 732 & \textbf{43.0} & 1,305 & \textbf{18.9} & 1,454 & \textbf{56.3} & \textbf{958} \\
\midrule
\multicolumn{11}{c}{\textit{Base: DeepScaleR-Preview-1.5B}} \\
\midrule
Baseline & 78.6 & 270 & 74.4 & 1,037 & 37.2 & 1,963 & 16.7 & 4,733 & 51.7 & 2,001 \\
L1-Max (512) & \underline{85.7} & 331 & 81.4 & 609 & 42.0 & 861 & 7.8 & 996 & 54.2 & \textbf{699} \\
L1-Max (1024) & \textbf{87.6} & 1,188 & \underline{82.2} & 1,235 & \underline{45.4} & 1,518 & \underline{22.2} & 1,661 & 59.4 & 1,401 \\
\textbf{HBPO (Ours)} & 85.6 & 394 & \textbf{82.4} & 726 & \textbf{47.2} & 1,193 & \textbf{22.2} & 1,476 & \textbf{59.4} & \underline{947} \\
\bottomrule
\end{tabular}
\caption{Performance under efficiency prompting setting. HBPO demonstrates robust performance compared to baseline models and the explicit length-controlled method L1, while effectively adhering to efficient prompting instructions.}
\label{tab:constrained_reasoning}
\end{table*}

\paragraph{Adaptive behavior emerges from hierarchical training rather than explicit control.} The distinction between HBPO and existing methods becomes evident in their token allocation patterns. L1-Max exhibits remarkably uniform behavior across problem difficulties, using 3,260 tokens on MATH500 and 3,163 tokens on AIME25 despite the significant complexity gap between these benchmarks. In contrast, HBPO demonstrates genuine problem sensitivity with token usage varying from 1,818 on MATH500 to 3,988 on AIME25. This 2.2× variation directly correlates with problem complexity and emerges naturally from the differentiated reward mechanism, which creates distinct optimization landscapes for different budget levels. Through comparative advantage across these landscapes, models learn to assess problem requirements without external guidance.

\section{Analysis}

\subsection{Analysis of Hierarchical Structure}

\begin{table*}[h!]
\centering
\small
\small
\begin{tabular}{lC{0.6cm}C{0.6cm}C{0.6cm}C{0.6cm}C{0.6cm}C{0.6cm}C{0.6cm}C{0.6cm}C{0.6cm}C{0.6cm}}
\toprule
\multirow{2}{*}{\textbf{Configuration}} & \multicolumn{2}{c}{\textbf{GSM8K}} & \multicolumn{2}{c}{\textbf{Math500}} & \multicolumn{2}{c}{\textbf{Olympiad}} & \multicolumn{2}{c}{\textbf{AIME25}} & \multicolumn{2}{c}{\textbf{Average}} \\
\cmidrule(lr){2-3} \cmidrule(lr){4-5} \cmidrule(lr){6-7}\cmidrule(lr){8-9} \cmidrule(lr){10-11} 
& Acc & Tokens & Acc & Tokens & Acc & Tokens & Acc & Tokens & Acc & Tokens \\
\midrule
Single ($b$=1536) & 85.6 & 327 & 83.4 & 1,055 & 48.1 & 2,301 & 22.2 & 3,686 & 59.8 & 1,842 \\
Dual ($b\in\{512,2560\}$) & 86.4 & 816 & 85.6 & 1,849 & 48.2 & 2,938 & 27.8 & 4,104 & 61.7 & 2,427 \\
\textbf{4-budget} & \textbf{87.6} & 790 & 86.2 & 1,818 & 50.0 & 2,861 & \textbf{31.1} & 3,988 & \textbf{63.7} & 2,364 \\
6-budget & 87.0 & 809 & \textbf{87.2} & 1,893 & \textbf{50.9} & 3,084 & 26.7 & 3,934 & 62.9 & 2,430 \\
8-budget & 87.4 & 864 & 85.6 & 1,836 & 49.9 & 2,899 & 28.9 & 4,019 & 62.9 & 2,405 \\
\bottomrule
\end{tabular}
\caption{Impact of hierarchical granularity on performance. The 4-budget configuration achieves optimal balance between and within-group learning and exploration diversity.}
\label{tab:budget_analysis}
\end{table*}

\paragraph{Optimal hierarchy emerges from balancing intra-group learning and inter-group exploration.}
To understand the impact of hierarchical structure on performance, we systematically analyze different budget configurations while maintaining a constant average budget of 1,536 tokens. Table~\ref{tab:budget_analysis} reveals a clear performance progression: single-budget training achieves only 59.8\% average accuracy, demonstrating the limitations of uniform exploration. The performance improves to 61.7\% with dual budgets and reaches an optimal of 63.7\% with our 4-budget configuration.

Single-budget training reduces to traditional uniform sampling without inter-budget reward differentiation. Dual budgets introduce basic differentiation between short (512) and long (2,560) reasoning, improving accuracy by 1.9\%. The 4-budget configuration achieves optimal performance by offering sufficient granularity for adaptive learning, while ensuring enough samples per subgroup to support effective intra-group optimization. Further increasing the number of budgets to 6 or 8 slightly degrades performance, with a 0.8\% drop, as fewer samples per subgroup weaken intra-group learning signals. This reveals a fundamental trade-off: exploration diversity must be balanced with statistical reliability for effective policy learning.

\paragraph{HBPO achieves efficiency through adaptive resource allocation rather than uniform compression.} As results shown in Table~\ref{tab:normal-performance}, traditional GRPO with cosine reward achieves some efficiency (average 1,150 tokens) but suffers significant accuracy degradation, particularly on complex tasks where it achieves only 23.3\% on AIME25. The model learns to generate universally short responses regardless of problem requirements, a form of mode collapse that sacrifices capability for efficiency.

\begin{table}[h!]
\centering
\caption{Comparison with traditional efficient reasoning methods under natural inference conditions.}
\label{tab:normal-performance}
\small
\begin{tabular}{lcccccccc}
\toprule
\multirow{2}{*}{\textbf{Method}} & \multicolumn{2}{c}{\textbf{GSM8K}} & \multicolumn{2}{c}{\textbf{MATH500}} & \multicolumn{2}{c}{\textbf{Olympiad}} & \multicolumn{2}{c}{\textbf{AIME25}}\\
\cmidrule(lr){2-3} \cmidrule(lr){4-5} \cmidrule(lr){6-7}\cmidrule(lr){8-9} 
& Acc & Tokens & Acc & Tokens & Acc & Tokens & Acc & Tokens \\
\midrule
Classic Reward & 86.2 & 661 & 86.2 & 1,605 & 49.1 & 3,174 & 24.4 & 4,309 \\
Cosine Reward & 83.0 & 195 & 77.6 & 478 & 42.0 & 1,271 & 23.3 & 2,657 \\
\textbf{HBPO}(Budget-aware Reward) & \textbf{87.6} & 790 & 86.2 & 1,818 & \textbf{50.0} & 2,861 & \textbf{31.1} & 3,988 \\
\bottomrule
\end{tabular}
\end{table}

\begin{figure*}[t]
  \centering
  \includegraphics[width=\columnwidth]{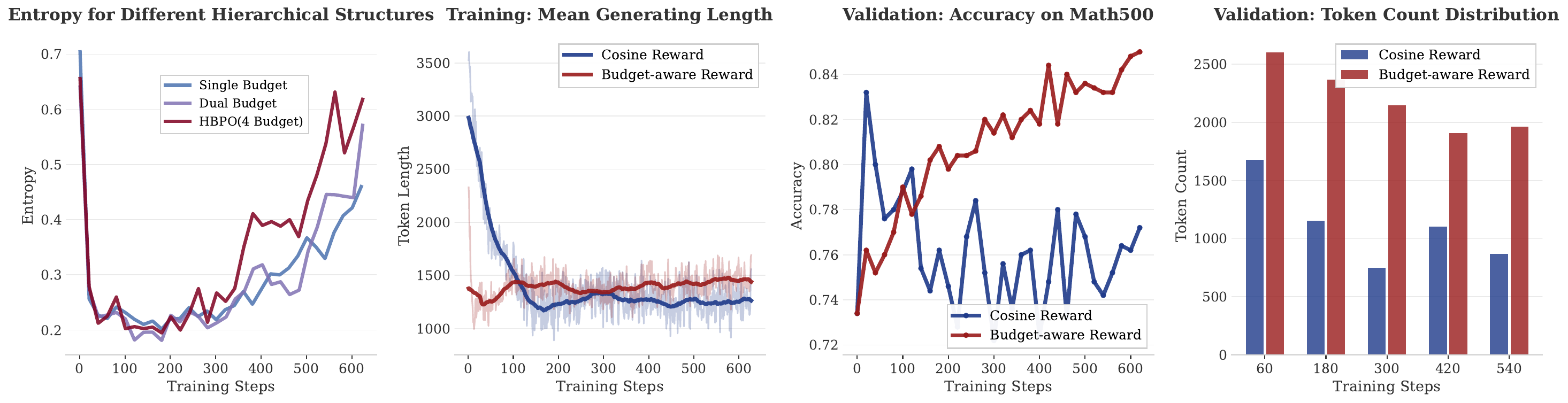}
  \caption{Training dynamics. (Left) Entropy Comparison of different hierarchical structures. (Right) Comparison of training dynamics and validation performance between cosine and budget-aware reward methods.}
  \label{fig:5-2}
\end{figure*}

Figure~\ref{fig:5-2} presents the training dynamics of entropy, mean generating length, and validation on the Math500 dataset, highlighting the advantages of hierarchical structures and budget-aware reward mechanism. HBPO (4-budget) setting significantly increases entropy throughout training, outperforming both the dual-budget and single-budget baselines. This suggests that a more fine-grained budget hierarchy encourages more diverse and effective exploration, thereby preventing exploration collapse. When comparing cosine reward to HBPO(budget-aware reward), the cosine reward leads to a sharp drop in generation length during the early training stages (steps 0–100), which results in excessive compression and poor generalization on the Math500 validation set. In contrast, HBPO maintains a stable average generation length of approximately 1,400 tokens. This stability stems from its hierarchical structure, which encourages effective exploration through budget-aware rewards rather than uniform compression. As a result, the model gradually discovers the most efficient reasoning length on the Math500 validation set during training and consistently improves its validation accuracy.

\subsection{Reasoning Pattern Analysis}

\begin{figure}[htbp]
    \centering
    \includegraphics[width=\textwidth]{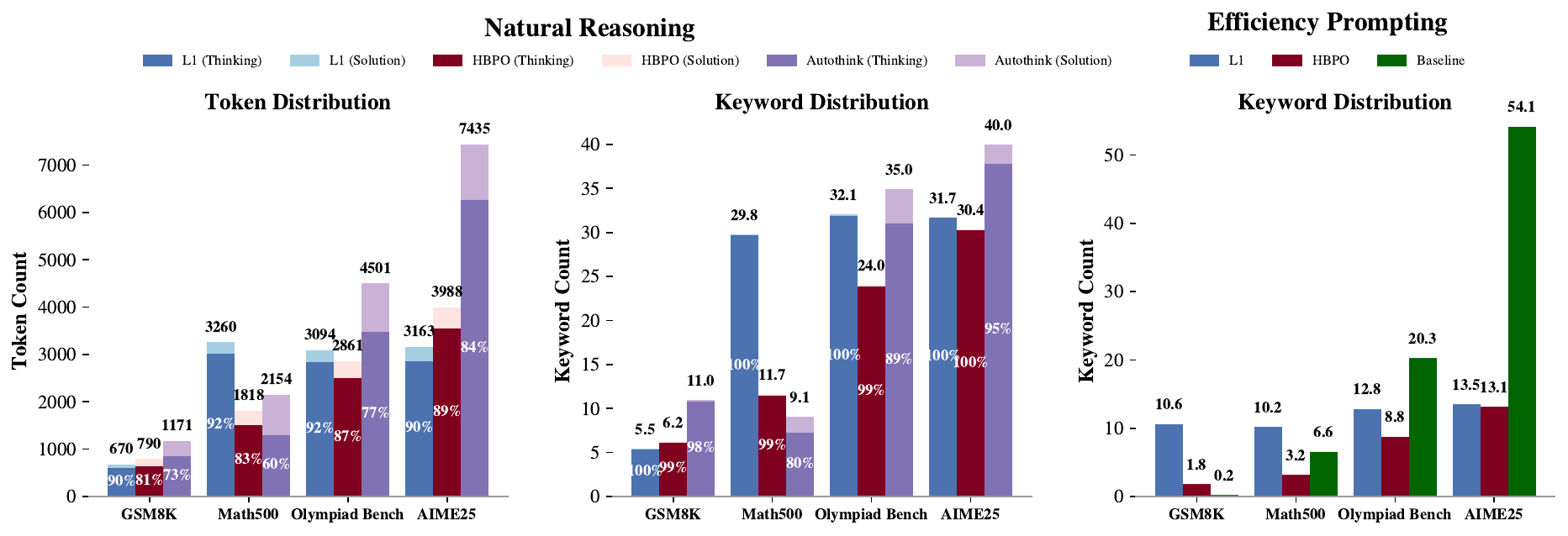}
    \caption{Reasoning pattern analysis across methods and problem difficulties. Thinking proportions and reflection keyword frequencies show HBPO's adaptive adjustment, with keywords properly contained within thinking segments.}
    \label{fig:5-3}
\end{figure}

\paragraph{HBPO develops different reasoning strategies based on problem complexity.}
To understand how models improve efficiency, we analyze reasoning patterns through two lenses: the proportion of exploratory thinking versus direct solution generation, and the frequency of reflection keywords that indicate deliberative processes. Figure~\ref{fig:5-3} reveals striking differences between methods.

HBPO exhibits clear adaptation to problem difficulty. The proportion of thinking content increases monotonically from 81\% on GSM8K to 89\% on AIME25, while reflection keywords (wait, alternatively, but, remember, check, and verify) rise from 6 to 30 occurrences per problem. This pattern supports our differentiated reward design, showing that the model learns to identify when longer reasoning adds value.

L1-Max improves efficiency through uniform length control, maintaining nearly constant thinking proportions (90-92\%) and keyword frequencies (29-32) across three datasets. This rigidity reveals mechanical optimization rather than intelligent adaptation. AutoThink attempts adaptive reasoning but exhibits problematic patterns: excessive thinking on simple problems (1171 tokens on GSM8K) and insufficient adjustment for complex ones. Moreover, AutoThink exhibits an average of 1.8 and 4.0 reasoning-related keywords per problem in the solution segments on the MATH500 and Olympiad benchmarks, indicating that reasoning processes leak into what should be direct answers.

The efficiency prompting setting provides further insight into adaptive capabilities. When instructed to minimize tokens, HBPO exhibits progressive keyword scaling (1.8 on GSM8K to 13.1 on AIME25), demonstrating that the model has internalized problem-complexity relationships. L1-Max, when explicitly prompted to ``think for 1024 tokens", shows minimal variation (10.6 to 13.5), revealing its inability to differentiate between problem requirements even under explicit efficiency instructions. These patterns confirm that hierarchical training enables genuine adaptive reasoning rather than uniform optimization.

\begin{wraptable}{r}{0.44\textwidth}
\centering
\caption{Performance on GPQA-Diamond}
\label{tab:ood}
\small
\begin{tabular}{lC{1.2cm}C{1.2cm}}
\toprule
\textbf{Model} & \textbf{Acc} & \textbf{Tokens} \\
\midrule
DeepScaleR & 33.84 & 4,762 \\
L1-Max & 33.33 & 1,227 \\
AutoThink & 34.41 & 3,787 \\
\textbf{HBPO} & \textbf{34.72} & 2,101 \\
\bottomrule
\end{tabular}
\end{wraptable}

\paragraph{Generalization to scientific reasoning validates domain-agnostic efficiency learning.}
To assess whether hierarchical exploration enables general efficiency principles rather than task-specific optimization, we evaluate on GPQA-Diamond, a challenging scientific reasoning benchmark outside our training domain. Table~\ref{tab:ood} shows that HBPO maintains the highest accuracy (34.72\%) while reducing token usage by 55\% compared to baseline. This performance on out-of-distribution tasks demonstrates that hierarchical training teaches fundamental principles of computational resource allocation that transfer across reasoning domains.

These analyses collectively demonstrate that HBPO's hierarchical exploration framework addresses the fundamental challenges in efficient reasoning. By maintaining exploration diversity through budget hierarchies and enabling adaptive learning through differentiated rewards, HBPO teaches models to recognize the computational requirements of different problems and allocate resources accordingly. The result is a system that achieves efficiency not through constraint but through understanding.

\section{Conclusion}

We introduced Hierarchical Budget Policy Optimization, a framework that enables reasoning models to achieve efficient computation without sacrificing capability. By maintaining diverse exploration through budget-constrained hierarchies and budget-aware rewards, HBPO prevents the exploration collapse and an optimized allocation of the length budget. Our experiments demonstrate that models trained with HBPO significantly reduce inference costs while improving performance, exhibiting adaptive behavior that naturally matches computational effort to problem complexity.

\bibliographystyle{iclr2026_conference}
\bibliography{ref}

\end{document}